%% file: main.tex
\documentclass{article}

\PassOptionsToPackage{numbers}{natbib}

 \usepackage[main, final]{neurips_2026}

\usepackage[utf8]{inputenc} 
\usepackage[T1]{fontenc}    
\usepackage{hyperref}       
\usepackage{url}            
\usepackage{graphicx}       
\usepackage{wrapfig}        
\usepackage{booktabs}       
\usepackage{multirow}       
\usepackage{amsfonts}       
\usepackage{nicefrac}       
\usepackage{microtype}      
\usepackage{xcolor}         
\usepackage{algorithm}
\usepackage{algorithmic}
\usepackage{enumitem}
\usepackage{caption}
 \usepackage{amsmath}

\newcommand{\model}{NeoWorld-Pro}

\title{NeoWorld-Pro: Programming Interactive Scenes from Monocular Images for Embodied Simulation}

\author{
\textbf{Yumeng~He}$^{1,2}$\thanks{Equal contribution.}
\qquad
\textbf{Eachen~Soong}$^{1,2}$\footnotemark[1]
\qquad
\textbf{Xiaotian~Yang}$^{3}$
\qquad
\textbf{Weijia~Zhang}$^{1}$
\\
\textbf{Zanwei~Zhou}$^{1}$
\qquad
\textbf{Junru~Gong}$^{1,2}$
\qquad
\textbf{Xiaokang~Yang}$^{1}$
\qquad
\textbf{Yunbo~Wang}$^{1,2}$\thanks{Corresponding author.}
\\
$^{1}$Shanghai Jiao Tong University 
\quad
$^{2}$Doxi AI
\quad
$^{3}$Huazhong University of Science and Technology
\\
\url{https://neoworldproject.github.io/neoworld-pro-website/}
}

\begin{document}

\maketitle
\enlargethispage{16pt}

\input{sec/0_abstract}    
\input{sec/1_intro}
\input{sec/2_related}
\input{sec/3_method}

\input{sec/4_results}
\input{sec/5_conclusion}




{
\small
\bibliographystyle{plain}
\bibliography{bib}
}





\end{document}

%% file: sec/0_abstract.tex
\begin{abstract}

The advancement of Embodied AI necessitates high-quality simulation assets that faithfully mirror the real world. However, transforming raw visual observations into simulation-ready scenes remains challenging due to the lack of physical grounding and scene-level interactivity in current image-to-URDF methods. We propose \textbf{NeoWorld-Pro}, a framework that reformulates monocular scene reconstruction as procedural programming for interactive 3D environments. Leveraging the zero-shot reasoning and code synthesis capabilities of MLLMs, NeoWorld-Pro converts a single RGB image into executable programs specifying object geometry, articulation, and physical properties. A \textbf{physics-in-the-loop} mechanism then iteratively refines the generated programs by validating their execution in a physics engine, enforcing physically plausible articulations, valid object compositions and interactions, and accurate spatial relationships. Experiments show that NeoWorld-Pro outperforms open-loop and prior monocular reconstruction methods, while enabling complex downstream tasks such as stable stacking and fine-grained manipulation.

\end{abstract}

%% file: sec/1_intro.tex
\section{Introduction}
\label{sec:intro}

The rapid advancement of embodied AI and robotics has increased the demand for high-quality, interactive simulation environments for robot learning.
In particular, building faithful digital twins requires not only accurate 3D geometry, but also correct kinematic structures, physical properties, and simulation-ready inter-object relationships that support \textit{Real-to-Sim-to-Real transfer}.
However, manually constructing such assets is labor-intensive and does not scale to the diversity and complexity of the real world.
As a result, automating the Real-to-Sim transition, from raw visual observations to physically functional, interactive simulation scenes, has become a critical research frontier.


Despite its importance, this transition remains constrained by three primary bottlenecks. First, most existing work focuses on \textit{single-asset image-to-URDF reconstruction}~\citep{chen2024urdformer,liu2024singapo,mandi2024real2code,le2024articulate,cao2025physx,wu2026urdf}, lacking scene-level modeling that jointly captures embodied interaction and inter-object spatial and physical dependencies.
Second, current approaches typically rely on large curated asset libraries or rigid category-level priors~\citep{chen2024urdformer,liu2024singapo,he2025spark,lu2025dreamart,li2026monoart,cao2025physx,wu2026urdf}, which imposes high data requirements and hinders generalization to \textit{out-of-distribution (OOD)} scenarios. 
Third, and perhaps most critically, existing pipelines are inherently \textit{open-loop}~\citep{chen2024urdformer,wu2026urdf,li2025urdf,he2025spark,cao2025physx}.
By treating reconstruction as a one-way inference task, they lack a mechanism for corrective feedback to resolve physical inconsistencies in geometry, articulation, or scene composition.
As a result, previous methods often struggle to produce environments that are truly ``simulation-ready'' for high-precision downstream tasks. 
Recent work, NeoWorld~\citep{zhao2025neoworld}, constructs explorable virtual worlds from a single image via progressive 3D unfolding, combining object-centric 3D foregrounds with 2D background synthesis for efficient exploration.
However, its hybrid neural representation offers limited explicit control over simulation-critical geometry, articulation, and physical properties.
Taken together, these limitations call for an explicit, executable world representation that supports structured editing and physics-based refinement.

\begin{figure}[t]
    \centering
    \includegraphics[width=\linewidth]{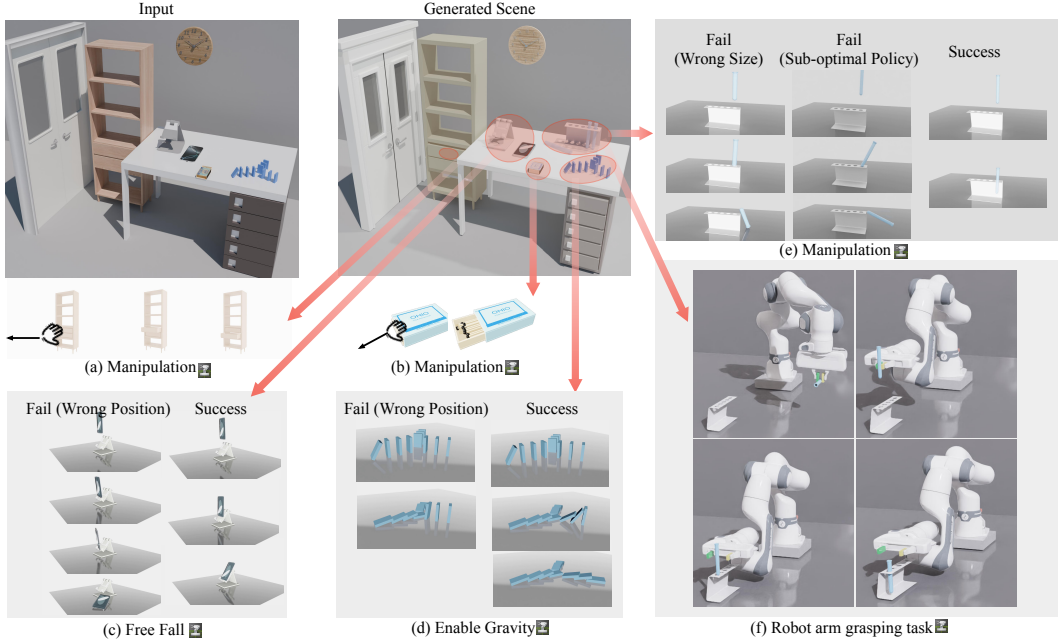}
    \vspace{-4pt}
    \caption{\textbf{Demonstration of NeoWorld-Pro.} Our framework ensures simulation-readiness by optimizing:
    (a-b) physically plausible intra-object articulations under external forces,
    (c-d) stable inter-object spatial configurations,
    and (e-f) scene viability for assembly and manipulation tasks.
    }
    \vspace{-10pt}
    \label{fig:scene_simulation}
\end{figure}

We propose NeoWorld-Pro---where ``Pro'' denotes \textit{programming}---a framework that reformulates monocular interactive scene construction through two core innovations: \textit{(1) asset reconstruction via procedural code synthesis} and \textit{(2) physics-in-the-loop scene refinement}.
In contrast to traditional dense reconstruction methods, which represent scenes as unstructured geometric entities like point clouds or meshes and prioritize per-pixel alignment, NeoWorld-Pro introduces a \textit{program-centric paradigm} that prioritizes functional and physical plausibility.
By leveraging the zero-shot spatial reasoning and code synthesis capabilities of Multimodal Large Language Models (MLLMs), our framework translates a monocular image into executable programs.
These programs explicitly define object geometry via compositional primitives, articulation parameters, physical properties, and scene layout.
Unlike static geometric representations, this program-centric approach is executable and editable, exposing the generated scene in a form that can be directly inspected, simulated, and modified.
By relying on the rich commonsense priors of MLLMs rather than dataset-specific training, NeoWorld-Pro generalizes effectively to novel categories and logically infers hidden structures, significantly improving robustness to the severe occlusions prevalent in cluttered scenes.

Furthermore, to upgrade superficial plausibility in the static appearance of generated scenes into true physical physical consistency, we introduce a \emph{physics-in-the-loop} validation and refinement mechanism. 
Instead of passively outputting static assets, \model{} executes the generated programs within a physics engine (\textit{e.g.}, Isaac Sim) to collect dynamic feedback, such as collision events, instability, and articulation failures.
This feedback is then fed back to the MLLM to iteratively refine the code.
As shown in Figure~\ref{fig:scene_simulation}, this closed-loop process validates joint executability at the object level while correcting inter-object penetration, unstable arrangements, and inaccurate relative scales at the scene level, ensuring the final environment is truly simulation-ready.

We evaluate \model{} on asset reconstruction, articulation prediction, scene assembly, and manipulation task execution.
Our results show that, by explicitly optimizing for simulation readiness through physics-in-the-loop refinement, \model{} significantly outperforms existing open-loop methods in generating environments that are immediately usable for robot learning.

\vspace{10pt}
In summary, the main contributions of this work are as follows:
\begin{itemize}[leftmargin=*]
    \vspace{-4pt}
    \item We introduce \model{}, formulating monocular scene construction as MLLM-driven program synthesis, reducing reliance on specific asset databases and supervised category-level priors.
    \vspace{-2pt}\item We propose a physics-in-the-loop scene refinement pipeline that leverages simulation feedback to iteratively improve both intra-object articulations and inter-object relationships.
    \vspace{-2pt}\item We construct a simulation benchmark comprising $90$ articulated object categories and $30$ scenes in Universal Scene Description (USD) format, and validate the effectiveness of \model{} on it.
    \vspace{-4pt}
\end{itemize}

%% file: sec/2_related.tex
\section{Related Work}
\label{sec:related}
\subsection{Articulated Asset Reconstruction}

The capability to model articulated assets is fundamental for enabling intelligent agents to perform complex manipulations in physical environments.
Yet, manually authoring such assets is labor-intensive, requiring explicit part decomposition, geometry modeling, and kinematic specification.
Recent learning-based methods thus aim to automate articulated asset construction from visual or geometric observations~\cite{li2020category,le2024articulate,lian2025infinite,geng2023gapartnet}.
Existing research predominantly relies on structured 3D inputs, such as static/dynamic point clouds~\cite{halacheva2025articulate3d,wang2026artllm,kreber2025guiding,yan2020rpm,weng2021captra,lin2025autourdf,yang2025artiworld} or meshes~\cite{zhang2026simart,xu2026motionanymesh,qiu2025articulate,huang2026react3d}, which are often unavailable in the wild. 
Alternative paradigms utilize multi-view-based 3D reconstruction~\cite{xia2025drawer,mandi2024real2code} or multi-state images~\cite{wu2025dipo,shen2025gaussianart,li2026gear,yuan2025larm,li2025art,chen2025freeart3d} to resolve kinematic ambiguity, yet incur significant data-acquisition overhead and limited scalability.
%

One line of work mitigates this ambiguity via retrieval- or assembly-based formulations, grounding symbolic structures in mesh libraries or templates~\cite{chen2024urdformer,liu2024singapo}. However, their generalization are constrained by the coverage and canonical biases of the underlying assets or training corpora.
Learning-based methods instead exploit category-level priors over part layouts, shapes, and motions~\cite{he2025spark,lu2025dreamart,li2026monoart}, reducing explicit retrieval dependence but remaining limited by the distributional coverage of training corpora and thus generalizing poorly to out-of-distribution structures or articulations.
More recent systems further target simulation-oriented asset construction by converting reconstructed geometry and inferred kinematics into executable representations for physics engines~\cite{wang2025kinematify,wu2026urdf,cao2025physx}. 

Although some of them evaluate generated assets in simulation~\cite{chen2024urdformer,wu2026urdf,li2025urdf} or downstream robotic tasks~\cite{le2024articulate,wang2025kinematify,cao2025physx3d,he2025spark}, such evaluation is typically post-hoc rather than used as feedback to refine the generation process. As a result, articulation and contact failures often emerge only after asset generation.
In contrast, our framework reformulates single-image asset construction as MLLM-driven procedural modeling and introduces physics-in-the-loop validation to refine both intra-object articulations and inter-object spatial consistency for simulation-ready scene generation.


\subsection{LLMs and MLLMs as 3D Generation Priors}
Recent studies have explored LLMs and MLLMs as priors for articulated 3D asset generation.
Some methods use foundation models mainly to construct supervision signals, textual conditions, physical annotations, or data augmentations~\cite{su2025artformer,cao2025physx,wu2026urdf}. 
Beyond such auxiliary usage, LLMs/MLLMs have been used as interpretive priors that expose the latent part structure and motion semantics of visual or geometric observations, while leaving the final asset construction to downstream retrieval, abstraction, or template-based modules~\cite{liu2024singapo,he2025spark,qiu2025articulate,wang2025kinematify,lu2025dreamart}.
SINGAPO~\cite{liu2024singapo} uses GPT-4o to infer an object-level part connectivity graph before grounding it through part abstraction and retrieval, while SPARK~\cite{he2025spark} leverages GPT-4o to parse parts and infer joints for constructing coarse URDF templates. Articulate-Anything~\cite{le2024articulate} further introduces an actor-critic formulation, where MLLM agents iteratively propose and critique link placement and joint configurations to improve articulation plausibility.

More recently, language models have been adapted from interpreters into direct generators of articulated asset specifications, producing structured outputs that can be converted into executable or simulator-compatible assets~\cite{mandi2024real2code,li2025urdf,yang2025artiworld,cao2025physx}. 
Among them, PhysX-Anything~\cite{cao2025physx} targets single-image physical asset generation by fine-tuning a MLLM to predict a unified structured representation of part geometry, articulation, and physical attributes.
Nevertheless, these methods are largely constrained by curated articulated-object or task-specific 3D and physical corpora, making their outputs inherit the structural and kinematic biases of the underlying datasetss.
In contrast, our framework reformulates single-view articulated asset generation as zero-shot code-driven procedural modeling, enabling more flexible generation beyond category-specific asset distributions.

%% file: sec/3_method.tex
\section{Method}
\label{sec:method}

The \model{} framework reformulates monocular scene reconstruction by shifting from unstructured geometric generation to a structured, program-centric paradigm. By representing scenes as compositions of executable programs, \model{} transforms visual scenes into simulation-ready functional entities.
In this section, we present the technical details of the \model{} framework.

\begin{figure}[t]
    \centering
    \includegraphics[width=\linewidth]{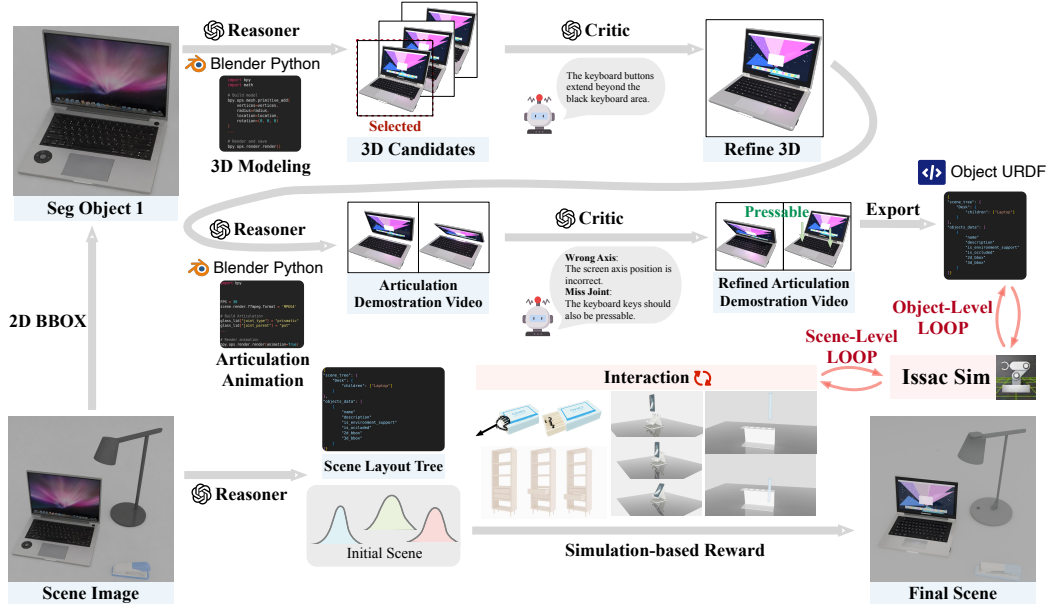}
    \vspace{-5pt}
    \caption{\textbf{Overview of the pipeline.} 
    Given a monocular RGB image, NeoWorld-Pro first performs scene parsing and reasoning to establish a scene tree and layout. Each foreground object is then programmed into an executable Blender script to generate geometry and articulation. A two-level \textit{physics-in-the-loop} mechanism ensures simulation-readiness: (1) an \textit{object-level loop} for resting stability and joint executability, and (2) a CEM-based \textit{scene-level loop} that optimizes scene plausibility, such as relative object scale and placement, collision-free spatial consistency, and stability.
    }
    \label{fig:method-overview}
    \vspace{-5pt}
\end{figure}

\subsection{Overview: Scene Reconstruction as Physics-in-the-Loop Program Synthesis}
\label{sec:overview}
Given a single RGB image $I$, our goal is to generate a simulation-ready interactive scene, represented as a collection of articulated object URDFs assembled within a Universal Scene Description (USD) stage $S$. 
This requires the simultaneous inference of part-level geometry, kinematics, and spatial relationships. To ensure these components are mutually consistent and physically stable within a simulator, NeoWorld-Pro moves beyond traditional dense reconstruction toward a structured, program-centric representation that bridges the gap between perception and physical execution.


As shown in Figure~\ref{fig:method-overview}, \model{} first leverages MLLMs to synthesize executable Blender Python programs as a unified intermediate representation, which are subsequently compiled into standard URDF and USD assets.
This program-centric paradigm offers several key advantages: it is \textit{executable} for direct physical simulation, \textit{interpretable} for MLLM-based inspection and verification, and \textit{editable} for iterative scene refinement.
Moreover, by exploiting the rich commonsense priors of foundation models, \model{} can logically infer occluded internal structures and missing physical attributes without relying on dataset-specific category priors.

Once the executable representation is established, the remaining challenge is to ensure its physical validity and simulation readiness.
We address this through a \textit{two-level physics-in-the-loop} framework that decomposes the pipeline into object-level and scene-level refinement stages, each designed to handle different types of reconstruction errors:
\begin{itemize}[leftmargin=*]
    \vspace{-4pt}\item
    Object-level loop (Sec.~\ref{sec:recon}): Refines discrete structural and physical attributes, including kinematic topology, joint limits, and inertial properties, through iterative MLLM-guided code editing.
    \vspace{-2pt}\item 
    Scene-level loop (Sec.~\ref{sec:loop}): Optimizes continuous scene assembly variables, such as relative scale, orientation, and spatial offsets, using the Cross-Entropy Method (CEM) for derivative-free optimization under simulation feedback.
    \vspace{-4pt}
\end{itemize}


\subsection{Object-Level Procedural Programming and Articulation Refinement}
\label{sec:recon}
This section describes how \model{} reconstructs physically consistent articulated objects and the corresponding scene layout from a single RGB image. The pipeline consists of three stages: scene parsing, procedural synthesis, and object-level physics refinement. Each stage follows a lightweight generate-critic-refine paradigm to improve robustness and physical consistency.

\vspace{-5pt}
\paragraph{Step 1: Scene parsing.} 
An \textbf{\texttt{MLLM-Reasoner}} (\textit{e.g.}, GPT-5.6-sol) first predicts (i) the scene hierarchy tree, (ii) 2D and 3D bounding boxes, and (iii) semantic attributes indicating whether an object serves as environmental support or is partially occluded.
This joint prediction allows relational information (\textit{e.g., a cup resting on a table}) to directly constrain object localization and layout estimation, avoiding heuristic downstream assembly. To improve parsing quality, an \textbf{\texttt{MLLM-Critic}} evaluates the predictions based on bounding-box tightness, completeness, and hierarchical consistency.
%
The resulting object crops are then forwarded to the procedural synthesis stage.

\vspace{-5pt}
\paragraph{Step 2: Procedural asset programming.} 
For each segmented object crop, an \textbf{\texttt{MLLM-Programmer}} sequentially generates (i) a Blender geometry program, (ii) an articulation program, and (iii) a URDF export script.
Each stage is wrapped within a runtime debugging loop to resolve execution failures, followed by a critic-refine loop in which rendered outputs are compared against the input crop.
We deliberately choose executable code over direct mesh generation as our intermediate representation, as it is highly reusable across stages and naturally leverages the MLLM's commonsense reasoning to hallucinate occluded internal structures (\textit{e.g., inferring a complete drawer cavity from a visible handle}) without relying on dataset-specific priors. 
The resulting URDFs are therefore structurally valid and articulation-aware, but their physical behavior remains unverified.

\begin{algorithm}[t]
\caption{CEM for Scene Composition Refinement}
\label{alg:scene-cem}
\begin{algorithmic}[1]
\STATE \textbf{Inputs:} Initial layout $\mu^{(0)}$, covariance $\Sigma^{(0)}$, iterations $T$, samples $N$, elite ratio $\rho$, input image $I$
\FOR{$t = 0 \dots T-1$}
    \STATE Sample $N$ candidates $\{z^{(n)}\}_{n=1}^N \sim \mathcal{N}(\mu^{(t)}, \Sigma^{(t)})$
    \FOR{$n = 1 \dots N$}
        \STATE Compose USD stage using base layout perturbed by $z^{(n)}$
        \STATE Run an Isaac Sim simulation rollout
        \STATE Compute physical penalties $D_{\mathrm{pen}}(z^{(n)})$ and $D_{\mathrm{drift}}(z^{(n)})$
        \STATE Render stage from canonical view and query MLLM for semantic score $S_{\mathrm{sem}}(z^{(n)}; I)$
        \STATE Compute total reward $R(z^{(n)})$
    \ENDFOR
    \STATE Select top-$K$ candidates where $K = \rho N$
    \STATE Update $\mu^{(t+1)}, \Sigma^{(t+1)}$ using the sample mean and variance of the top-$K$ elites
    \IF{convergence criteria met}
        \STATE \textbf{break}
    \ENDIF
\ENDFOR
\STATE \textbf{Return:} Best candidate layout
\end{algorithmic}
\end{algorithm}

\begin{figure}[!t]
    \centering
    \includegraphics[width=\linewidth]{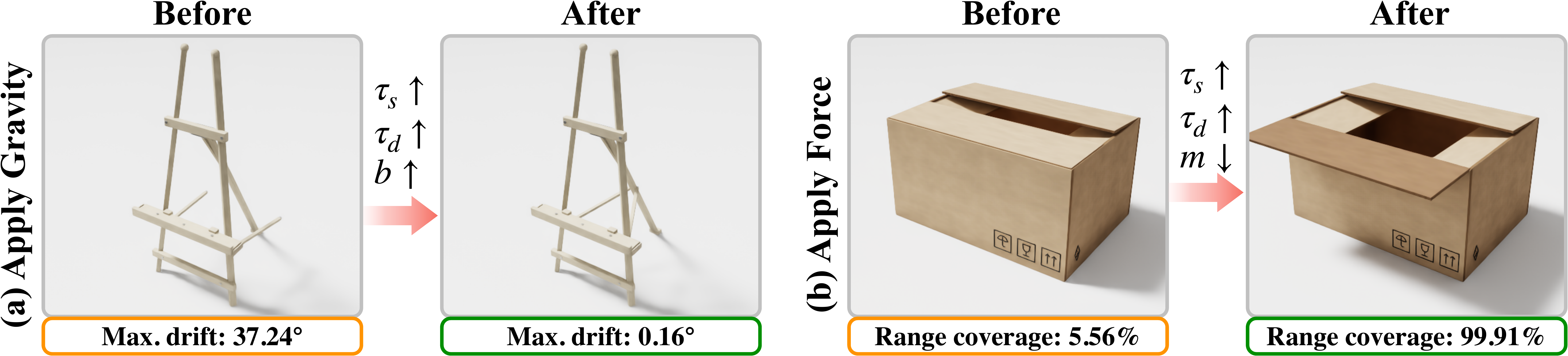}
    \vspace{-10pt}
    \caption{
\textbf{Physics-in-the-loop refinement improves resting stability and joint executability.}
(a) Increasing static friction torque $\tau_s$, dynamic friction torque $\tau_d$, and passive viscous damping $b$ reduces the maximum angular drift of the easel braces under gravity from $37.24^\circ$ to $0.16^\circ$.
(b) Reducing $\tau_s$, $\tau_d$, and flap mass $m$ increases joint range coverage from $5.56\%$ to $99.91\%$ under force limit.
}
    \label{fig:physics}
    \vspace{-4pt}
\end{figure}

\vspace{-5pt}
\paragraph{Step 3: Object-level physics-in-the-loop refinement.}
To ensure physical validity, the generated URDFs are evaluated in Isaac Sim under two standardized physics scenarios: \textit{resting stability} and \textit{joint executability}.
We use CoACD~\cite{wei2022approximate} to generate collision geometries for the reconstructed objects.
The stability test assesses whether each object maintains its intended resting configuration under gravity, without toppling, penetration, or unintended disassembly.
For articulated objects, actuation tests determine whether the intended joint motion can be completed under physically plausible force or torque limits.
Acceptance is determined by numerical criteria computed from simulator-recorded trajectories.
For failed tests, the \textbf{\texttt{MLLM-Reasoner}} integrates trajectory measurements with simulation frames to diagnose failure mechanisms and propose parameter updates, as illustrated in Fig.~\ref{fig:physics}.
%
%
%
The validated resting pose of each object subsequently serves as its initial configuration for scene-level composition.
At this stage, each reconstructed object is individually physically consistent. However, ensuring correct relative scale, placement, and interaction among objects requires scene-level optimization, which we introduce next.

\begin{figure}[t]
    \centering
    \includegraphics[width=\linewidth]{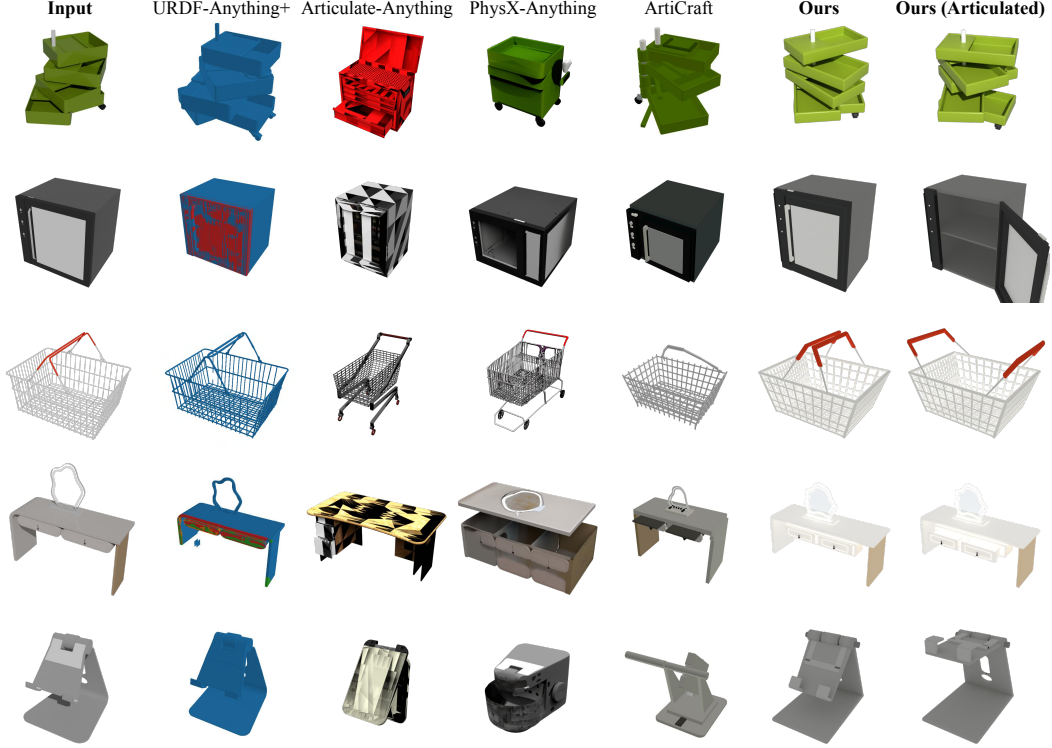}
    \vspace{-10pt}
    \caption{\textbf{Object-level reconstruction results of articulated assets.} Compared with existing methods, \model{} recovers more faithful geometry, appearance, and articulation structure.}
    \label{fig:object_recon}
    \vspace{-10pt}
\end{figure}

\begin{figure}[t]
    \centering
    \includegraphics[width=\linewidth]{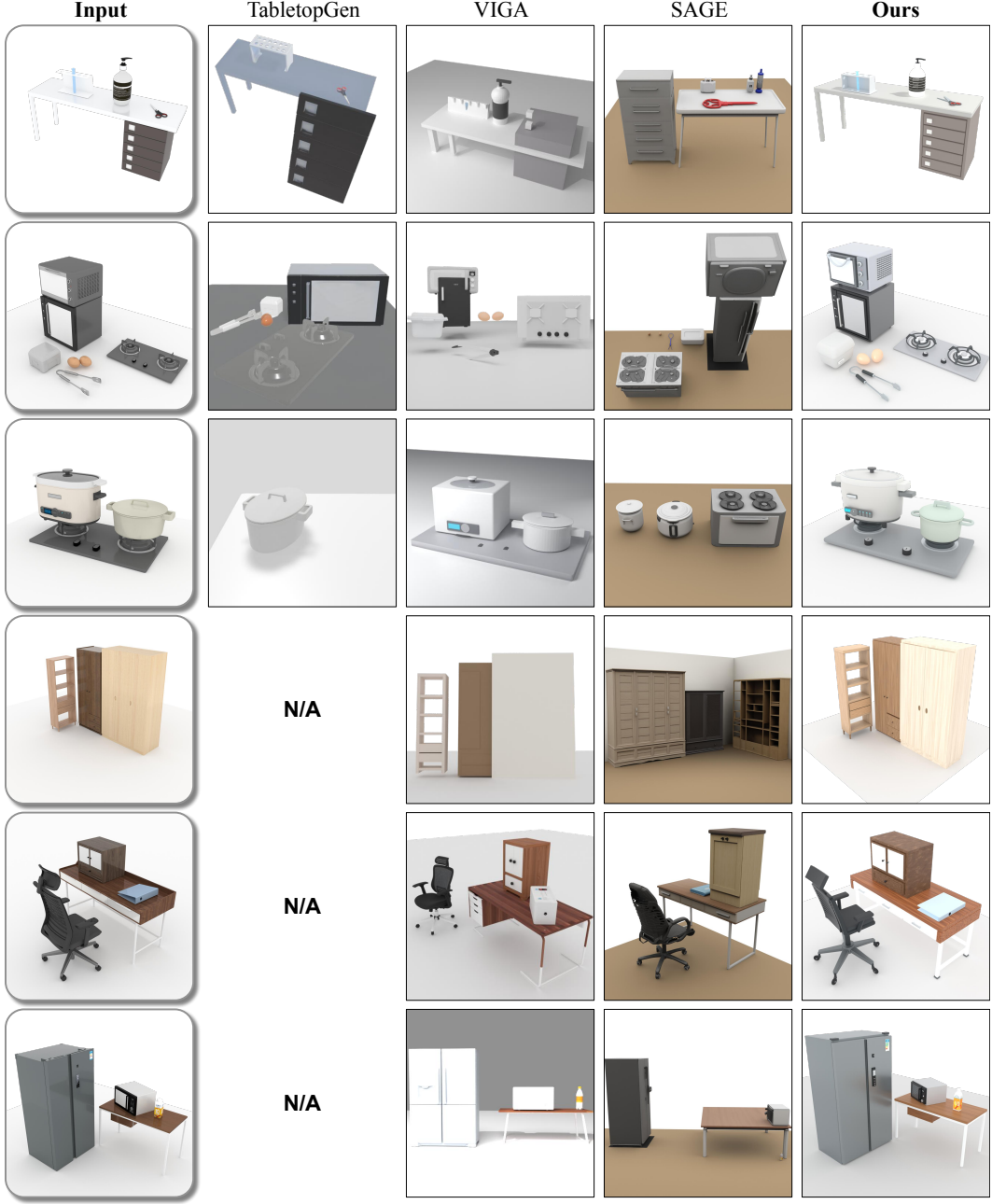}
    \vspace{-5pt}
    \caption{\textbf{Qualitative scene reconstruction on our synthetic scene dataset.}
    Each row compares reconstructions from the same monocular input. \model{} more faithfully preserves object identity, relative scale and orientation, and inter-object layout across both tabletop and larger workspace scenes. ``N/A'' denotes examples outside TabletopGen's tabletop setting.
    }
    \label{fig:synthetic_scene}
    \vspace{-5pt}
\end{figure}

\subsection{Scene-Level Composition and CEM Refinement}
\label{sec:loop}

This section assembles the generated URDF assets into a complete and refined USD, which consists of two stages: deterministic scene composition from the inferred layout, followed by stochastic optimization via a reward-driven CEM that integrates physical simulation and MLLM-based evaluation.

\vspace{-5pt}
\paragraph{Step 4: Scene composition.} 
We construct the initial scene via a top-down traversal of the scene tree, using each object's validated resting pose as its local configuration. The horizontal coordinates $(x,y)$ and yaw rotation are initialized from the parsed spatial layout, while the vertical coordinate is determined by aligning each child's transformed bottom surface with the top surface of its parent. This strategy provides a geometrically consistent initialization while reducing floating and inter-object penetration.
Non-environment objects are instantiated via USD references to preserve their URDF articulation properties, whereas environment supports are instantiated as static primitives and excluded from subsequent optimization.
Although this deterministic construction provides a plausible initialization, it often introduces physical artifacts such as inter-object penetration, unstable stacking, scale inconsistency, and incorrect relative orientation. These errors are difficult for MLLMs to correct in continuous pose space but can be directly quantified using a physics engine.

\begin{figure}[t]
    \centering
    \includegraphics[width=\linewidth]{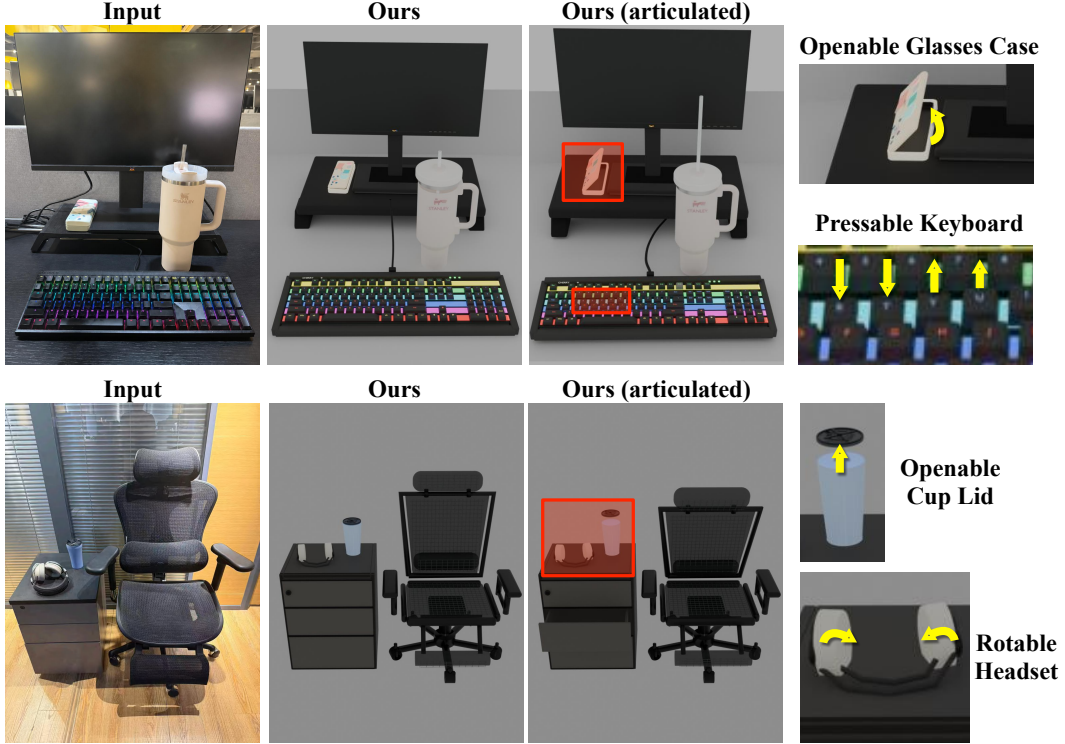}
    \vspace{-10pt}
    \caption{\textbf{Reconstruction from real-world images.}
    The results illustrate generalization to real-world appearance and clutter while retaining functional articulations.
    }
    \label{fig:real_scene}
    \vspace{-10pt}
\end{figure}

\vspace{-5pt}
\paragraph{Step 5: CEM-based scene refinement.}
We maintain a diagonal Gaussian distribution over a four-dimensional perturbation vector for each non-environment object $o_i$. We treat the initial layout as the zero-mean configuration, with $z_i = (\Delta x_i, \Delta y_i, \Delta\psi_i, \log s_i) \in \mathbb{R}^4$ representing axial translation, yaw rotation, and isotropic scale in parent coordinates.
We use blockwise CEM, selecting an active object group based on the
current failure relationships. Each iteration samples $N=16$ candidates
from the bounded Gaussian distributions using rejection sampling.
The top $K=4$ candidates determine the updated means and variances
of the active group. Standard deviations are bounded below by
$20\%$ of their initial values. If all candidates have equal rewards,
the distributions remain unchanged.
%
%
%
At each iteration, we sample $N$ candidate layouts, instantiate them in USD, and simulate them in Isaac Sim with gravity enabled to evaluate the stability of inter-object support relationships.
The top-$K$ candidates are updated under the Gaussian distribution via a hybrid reward, where $w_* \ge 0$ are scalar weights, $I$ is the input image, and $z$ is the candidate perturbation vector:
\begin{equation}
    R(z) = w_{\mathrm{sem}} S_{\mathrm{sem}}(z; I) - w_{\mathrm{pen}} D_{\mathrm{pen}}(z) - w_{\mathrm{drift}} D_{\mathrm{drift}}(z).
\end{equation}
Here, physical terms are measured directly in simulation: $D_{\mathrm{pen}}$ aggregates threshold exceedances of maximum PhysX contact penetration across object pairs, while $D_{\mathrm{drift}}$ combines orientation deviation, terminal pose and velocity variation, and support and attachment errors.
The semantic term $S_{\mathrm{sem}}$ uses an MLLM to assess rendered candidates against the input image, emphasizing relational cues such as depth ordering, relative scale, and spatial consistency.
This process repeats until convergence or a fixed iteration budget (Algorithm~\ref{alg:scene-cem}).
%
By resolving discrete structural attributes in Sec.~\ref{sec:recon} and continuous scene optimization in this section, \model{} forms a unified physics-in-the-loop framework that jointly enforces articulation correctness and global spatial consistency, distinguishing it from prior open-loop approaches.

%% file: sec/4_results.tex
\section{Experiments}
\label{sec:exp}

\subsection{Experimental Setups}

We evaluate \model{} on both a public articulated-object benchmark and a newly constructed multi-object simulation benchmark.
We introduce a synthetic scene dataset for monocular reconstruction in physically realistic multi-object environments.
The dataset contains $90$ articulated object categories, composing $30$ USD-format object assemblies in total.
Each scene contains multiple articulated or rigid objects with both intra-object and inter-object occlusions, and is paired with several downstream manipulation tasks for evaluation beyond reconstruction and articulation prediction.


For single-object image-to-URDF reconstruction, we compare \model{} with four representative baselines: 
PhysX-Anything~\cite{cao2025physx} and URDF-Anything+~\cite{wu2026urdf}, which generate simulation-compatible object assets from visual inputs; Articulate-Anything~\cite{le2024articulate}, which follows a retrieval-and-critic paradigm for articulated structure estimation; and Articraft~\cite{zhou2026articraft}, a state-of-the-art generation method that synthesizes articulated assets through agentic program synthesis, which mainly focuses on text input. For scene-level reconstruction, we compare against TabletopGen~\cite{wang2025tabletopgen}, VIGA~\cite{yin2026vision}, and SAGE~\cite{xia2026sage}, spanning simulation-ready generation from tabletop to room-scale scenes.

\begin{table}[t]
    \centering
    \small
    \setlength{\tabcolsep}{9pt}
    \caption{\textbf{Object-level evaluation of joint articulation and kinematics.} \textit{\#Total}, \textit{\#Pred}, and \textit{\#Hit} are the number of ground-truth joints, predicted joints, and matched hits. We report the joint miss rate (\textit{Miss}),  joint type identification error (\textit{Type}), orientation error (\textit{Axis}), and positional error (\textit{Pivot}).
    }
    \vspace{4pt}
    \label{tab:articulation_prediction}
    \resizebox{\linewidth}{!}{%
    \begin{tabular}{l ccccccc }
        \toprule
        Method  & Total & \#Pred & \#Hit & Miss $\downarrow$ & Type$\downarrow$ & Axis$\downarrow$ & Pivot$\downarrow$ \\
        \midrule
        Articulate-Anything~\cite{le2024articulate} &245&170&95&61.22 & 69.39 & 63.25 & 0.32 \\
        PhysX-Anything~\cite{cao2025physx}          &245&192&126&48.57 & 58.78 & 45.80 & 0.37 \\
        URDF-Anything+~\cite{wu2026urdf}            &245&96&74&69.80 & 82.86 & 50.41 & 0.40 \\
        Articraft~\cite{zhou2026articraft}          &245&304&190&22.45 & 29.39 & 30.14 & 0.23 \\
        \textbf{Ours}                               &\textbf{245}&\textbf{268}&\textbf{216}&\textbf{11.83} & \textbf{18.78} & \textbf{28.52} & \textbf{0.19} \\
        \bottomrule
    \end{tabular}
    }
     \vspace{-10pt}
\end{table}

\begin{table}[t]
    \centering
    \footnotesize
    \setlength{\tabcolsep}{4pt}
    \caption{\textbf{Scene-level results on the full benchmark and the 11-scene tabletop subset.} S-Pen, P-Pen, S-Stab, and O-Stab denote scene-level (inter-object), part-level (inner-object) penetration rate, and scene-/object-level stability, respectively. These metrics are percentages. MLLM scores (the last three columns) use a $1$--$7$ scale.}
    \label{tab:scene_quantitative}
    \vspace{3pt}
    \resizebox{\linewidth}{!}{%
    \begin{tabular}{lccccccccc}
        \toprule
        Method & S-Pen$\downarrow$ & P-Pen$\downarrow$ & S-Stab$\uparrow$ & O-Stab$\uparrow$ & CLIP$\uparrow$ & DINOv2$\uparrow$ & Complete$\uparrow$ & Layout$\uparrow$ & Function$\uparrow$ \\
        \midrule
        \multicolumn{10}{c}{\textit{Full benchmark}} \\
        VIGA~\cite{yin2026vision} & 56.67 & 63.37 & 23.33 & 27.33 & 0.8831 & 0.6087 & 4.70 & 3.95 & 3.45 \\
        SAGE~\cite{xia2026sage} & 6.67 & 2.41 & 70.00 & 91.60 & 0.8729 & 0.5379 & 3.80 & 2.60 & 2.90 \\
        \textbf{Ours} & \textbf{0.00} & \textbf{0.00} & \textbf{100.00} & \textbf{100.00} & \textbf{0.9128} & \textbf{0.7855} & \textbf{6.20} & \textbf{5.35} & \textbf{5.58} \\
        \midrule
        \multicolumn{10}{c}{\textit{Tabletop subset}} \\
        VIGA~\cite{yin2026vision} & 54.55 & 67.26 & 18.18 & 25.66 & 0.8764 & 0.5950 & 4.55 & 3.91 & 3.00 \\
        TabletopGen~\cite{wang2025tabletopgen} & 45.45 & 26.53 & 36.36 & 67.35 & 0.8917 & 0.6266 & 4.36 & 3.73 & 3.64 \\
        SAGE~\cite{xia2026sage} & 9.09 & 4.17 & \textbf{100.00} & \textbf{100.00} & 0.8741 & 0.4742 & 3.45 & 2.64 & 2.73 \\
        \textbf{Ours} & \textbf{0.00} & \textbf{0.00} & \textbf{100.00} & \textbf{100.00} & \textbf{0.9205} & \textbf{0.7166} & \textbf{6.10} & \textbf{5.70} & \textbf{5.58} \\
        \bottomrule
    \end{tabular}%
    }
     \vspace{-5pt}
\end{table}

\subsection{Object-Level Reconstruction Results of Articulated Assets}

We present qualitative results of single-object image-to-URDF reconstruction in Figure~\ref{fig:object_recon}. As shown, our method achieves the most accurate recovery of articulated joints and produces highly accurate geometric reconstructions.
Table~\ref{tab:articulation_prediction} reports quantitative articulation results. \model{} achieves the lowest joint miss rate, the largest number of matched joints, and the lowest joint type, axis, and pivot errors among the evaluated methods, indicating more complete and accurate joint recovery. Detailed evaluation protocols and additional results on appearance, geometry, and cross-modal similarity are provided in Appendix~A.1.



\subsection{Scene-Level Reconstruction and Assembly Results}

%


We provide a qualitative comparison on our synthetic scene dataset in Figure~\ref{fig:synthetic_scene}. 
Across tabletop, appliance, and workspace layouts, \model{} preserves the complete set of foreground objects while more faithfully recovering their relative scales, orientations, and spatial relationships. In contrast, competing methods often omit or replace objects and distort layouts, especially as the number and diversity of objects increase.
%
%
Beyond static reconstruction quality, we further demonstrate that our reconstructed scenes are free of object penetration, remain stable under dynamic simulation, and support physically meaningful downstream interactions, including stable phone–stand support, robotic grasping of test tubes from racks, and controllable domino-pushing behaviors.
These scenarios require accurate recovery of inter-object geometry, articulation, contact, and spatial dependencies, which existing open-loop methods struggle to achieve reliably.

For quantitative comparisons, Table~\ref{tab:scene_quantitative} summarizes scene-level inter-object penetration rate, part-level self-penetration rate, and scene- and object-level stability under gravity. 
All physical metrics are evaluated in a physics simulator (\textit{e.g.}, Isaac Sim).
%
%
We additionally evaluate CLIP and DINOv2 similarity, together with MLLM-based judgments of scene completeness, layout fidelity, and functional feasibility.
MLLM-based scores assess whether the reconstructed scene preserves all relevant objects (\emph{completeness}), faithfully reproduces their spatial arrangement (\emph{layout fidelity}), and supports the intended physical interactions (\emph{functional feasibility}).
Across both the full benchmark and the tabletop subset, \model{} eliminates inter-object collisions, maintains strong stability, and achieves the highest semantic scores. 
Figure~\ref{fig:real_scene} further demonstrates generalization to two real-world monocular images. Despite background clutter, illumination variation, and partial occlusion, \model{} reconstructs the primary objects and their spatial relationships. The actuated renderings additionally show that the inferred movable components remain functional after scene assembly.




\subsection{Image-to-Policy Demonstration for the Manipulation Task}

As illustrated in Figure~\ref{fig:scene_simulation}, we further demonstrate image-to-policy execution on a test-tube placement task.
Given a monocular scene image containing a test tube and a rack, \model{} reconstructs both objects and their relative spatial configuration.
Given the test-tube placement task, which requires the tube to pass through an opening and be supported by the rack, an MLLM specifies the initial end-effector position and displacement for executing the manipulation in Isaac Sim.
Errors in the relative pose or scale of the reconstructed objects may render the predicted motion infeasible and cause task failure.
We use this execution outcome as a feedback signal to iteratively refine the scene layout until the placement succeeds, demonstrating that the reconstructed scene supports physically grounded downstream manipulation rather than merely visual resemblance.

\subsection{Ablation Studies}
\noindent
\begin{minipage}[t]{0.53\linewidth}
We validate the refinement components of \model{} at both object and scene levels, evaluating their effects on articulation recovery and scene-level task execution. At the object level, we remove either the geometry or articulation critic to assess their impact on joint recovery. At the scene level, we removes closed-loop simulation feedback and directly uses the initially composed scene.
\end{minipage}
\hfill
\begin{minipage}[t]{0.45\linewidth}
\vspace{\dimexpr -15pt+\baselineskip\relax}
\centering
\captionof{table}{\textbf{Analyses of model component.}}
\label{tab:ablation}
\vspace{-6pt}
\setlength{\tabcolsep}{3pt}
\small
\begin{tabular}{lcc}
        \toprule
        Method                              & Success (\%) & Miss (\%) \\
        \midrule
        w/o geom. refine critic             & 76.47    & 14.84 \\ 
        w/o artic. refine critic            & 83.33    & 17.18 \\ 
        w/o simul. (open-loop)    & 44.05    & 11.72 \\ 
        \textbf{\model{}}                            & \textbf{92.85} & \textbf{7.03} \\ 
        \bottomrule
    \end{tabular}
    \vspace{-5pt}
\end{minipage}

As shown in Table~\ref{tab:ablation}, without physical validation, errors from parsing and generation remain uncorrected, leading to failures such as penetration, unstable stacking, and incorrect scale, as illustrated in Figure~\ref{fig:scene_simulation}.
%
Removing the semantic score, inter-object penetration penalty, or stability penalty reduces task success, indicating that all three terms contribute to reliable scene-level execution.

%% file: sec/5_conclusion.tex
\section{Conclusion and Limitation}
\label{sec:conclusion}

We presented \model{}, a physics-in-the-loop framework for monocular real-to-sim scene reconstruction. By reformulating reconstruction as MLLM-driven program synthesis and coupling it with object-level and scene-level simulation feedback, \model{} produces simulation-ready assets with accurate geometry, articulation, and spatial relationships. Extensive experiments demonstrate strong improvements over open-loop baselines in appearance fidelity, geometric accuracy, articulation prediction, and downstream manipulation tasks, highlighting the effectiveness of integrating executable representations with physical validation.

\model{} is currently limited to rigid and articulated objects under quasi-static physics. It does not yet model highly deformable or continuous media such as cloth, fluids, or granular materials, where dynamics are significantly more complex and less amenable to URDF-style representations. Our evaluation is also limited to scenes with fewer than $10$ objects. Extending the framework to support deformable assets, richer material properties, and denser scenes remains an important direction for future work. Improving computational efficiency to accelerate scene reconstruction is another important direction for future work.

\section*{Acknowledgements}
This work was supported by the National Natural Science Foundation of China (Grant 62250062), the Shanghai Municipal Science and Technology Major Project (Grant 2021SHZDZX0102), and the Fundamental Research Funds for the Central Universities. 

\section*{Competing-interests Disclosure}
The authors declare that they have no competing interests.